\documentclass[10pt,twocolumn,letterpaper]{article}

\usepackage{cvpr}
\usepackage{times}
\usepackage{epsfig}
\usepackage{times}

\usepackage{epsfig}
\usepackage{graphicx}
\usepackage{amsmath}
\usepackage{amssymb}
\usepackage{authblk}

\usepackage{tabularx}

\usepackage[utf8]{inputenc}
\usepackage[T1]{fontenc}
\usepackage{textcomp}
\usepackage{gensymb}
\usepackage{multirow}
\usepackage{booktabs}
\usepackage{xcolor}
\usepackage{graphicx}
\usepackage{float}
\usepackage{capt-of}
\usepackage{etoolbox}
\usepackage{adjustbox}
\usepackage[font=small]{caption}
\usepackage[ruled,vlined]{algorithm2e}

\usepackage{paralist}

\usepackage[font=small,skip=10pt]{caption}
\usepackage{float}
\usepackage{makecell}

\usepackage{wrapfig}

\newcommand{\mb}[1]{\mathbf{#1}}
\newcommand{\mcal}[1]{\mathcal{#1}}

\newcommand{\rrc}{\emph{RRB}}

\usepackage[breaklinks=true,bookmarks=false]{hyperref}

\makeatletter
\renewcommand\AB@affilsepx{, \protect\Affilfont}
\makeatother

\cvprfinalcopy 

\def\cvprPaperID{4} 

\ifcvprfinal\fi
\begin{document}

\title{Reconstruct, Rasterize and Backprop: Dense shape and pose estimation\\ from a single image}

\author[1]{Aniket Pokale\thanks{Authors contributed equally.}}
\author[1]{Aditya Aggarwal$^*$}
\author[2]{Krishna Murthy Jatavallabhula}
\author[1]{K. Madhava Krishna}
\affil[1]{Robotics Research Center, KCIS, IIIT Hyderabad, India}
\affil[2]{Mila, Universite de Montreal, Canada}

\pagenumbering{gobble}
\makeatletter
\let\@oldmaketitle\@maketitle
\renewcommand{\@maketitle}{\@oldmaketitle
\centering
\includegraphics[width=17cm]{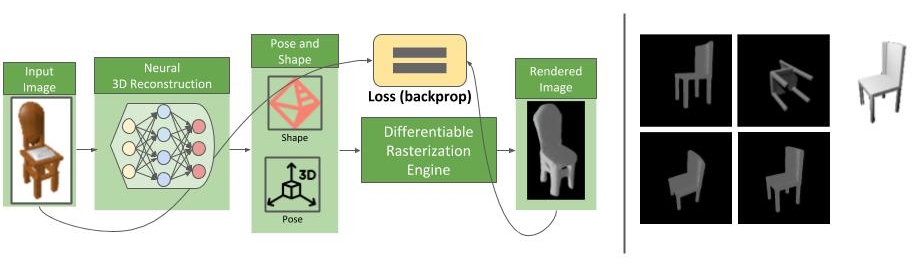}
\vspace{-0.5cm}
\captionof{figure}{
\textbf{Reconstruct, Rasterize and Backprop}: An input image is first passed through an \emph{occupancy network}~\cite{onet} to generate an output triangle mesh. Then, a pose estimation block employs a differentiable rasterization engine~\cite{kato2018neural} to refine the pose of this mesh, by comparing it with the input image. On the right we show intermediate poses of the optimization process, demonstrating that the renderer helps recover from extremely poor rotation initializations.}
\vspace{0.25cm}
\label{fig:teaser}
}
\makeatother

\maketitle
\begin{abstract}
\vspace{-0.25cm}
This paper presents a new system to obtain dense object reconstructions along with 6-DoF poses from a single image. Geared towards high fidelity reconstruction, several recent approaches leverage implicit surface representations and deep neural networks to estimate a 3D mesh of an object, given a single image. However, all such approaches recover only the \emph{shape} of an object; the reconstruction is often in a \emph{canonical frame}, unsuitable for downstream robotics tasks.
To this end, we leverage recent advances in differentiable rendering (in particular, rasterization) to \emph{close the loop} with 3D reconstruction in camera frame. 

We demonstrate that our approach---dubbed \emph{reconstruct, rasterize and backprop} (RRB)---achieves significantly lower pose estimation errors compared to prior art, and is able to recover dense object shapes and poses from imagery.
We further extend our results to an (offline) setup, where we demonstrate a dense monocular object-centric egomotion estimation system.
\end{abstract}

\vspace{-0.5cm}
\section{Introduction}
\label{sec:introduction}

As more robots continue to prevade our workplaces, homes, and lives, developing accurate and robust algorithms for their effective and safe operation is of paramount importance. Nearly all robots operating in domestic environments (workplaces and homes) leverage high quality maps in one form or the other. To be more interpretable, for both robots and humans, several recent approaches \cite{quadricslam,salas2013slam++,parkhiya2018constructing} have argued in favour of building feature-rich maps laden with semantics and object-level labels.

This line of thought has further been fueled with the advent of deep neural networks \cite{krizhevsky2012imagenet}, which have enabled tremendous strides in terms of robot perception. In particular, the otherwise ill-posed task of estimating poses and shapes of objects from a single image has been made possible with neural networks. Most approaches to this task either reconstruct objects as primitives (bounding boxes~\cite{cubeslam}, quadrics~\cite{quadricslam}), wireframes~\cite{krishna-icra2017}, or as a collection of features. A few other approaches \cite{variationalobjectslam,Toppe_2013_CVPR} estimate volumetric shapes of objects at a specified (coarse) resolution.

In 3D vision research, reconstruction quality has been bolstered by recent advances in leveraging \emph{implicit surface representations} to simplify the task at hand. In particular, signed distance functions have been effectively used to produce high quality 3D reconstructions from images \cite{onet,xu2019disn, saito2019pifu}. While the shapes estimated from these approaches are dense and of high fidelity, the notion of a \emph{camera pose} in such setups is ambiguous. 

None of the approaches that produce dense shapes, output object poses with respect to the camera. The other class of approaches \cite{parkhiya2018constructing,wang20196} that produce accurate poses do not recover dense object shapes. This is the key gap that we address in this paper. We propose a technique to recover dense shapes and poses of objects relative to a moving monocular camera.

We bring the best of both worlds by leveraging differentiable rasterization approaches \cite{kato2018neural,liusoft}. Specifically, we present \emph{reconstruct, rasterize and backprop} (\rrc), a system that marries deep neural networks and differentiable rasterization engines to build an accurate dense shape and pose estimator.

The key ingredients of the proposed framework include:
\begin{enumerate}
    \item An \emph{occupancy network}~\cite{onet} that produces implicit surface representations from a single image, in a \textit{canonical} object frame.
    \item A viewpoint initialization network, that \emph{guesses} a (noisy) transform from the canonical object frame to the camera coordinate frame.
    \item A differentiable rasterization engine that takes as input the initial shape and pose estimates and \emph{renders} images.
    \item Gradient-based optimization machinery (usually accelerated gradient methods) that compare the rendered image against a silhouette of the original image, and iteratively update the pose parameters until the rendered image matches the silhouette.
\end{enumerate}
    
Such a system has a number of potential advantages over existing single-image reconstruction frameworks. \rrc{} produces high fidelity (mesh-based) shapes and poses of objects from still images. Further, we demonstrate that this framework can be applied to several offline robotics tasks such as reconstruction, object-centric egomotion estimation, and data association.

\section{Related Work}
\label{sec:relatedwork}
In this section we discuss existing approaches to 3D reconstruction from the single images, 6D pose recovery, object-based SLAM and differentiable rendering.

\subsection{3D reconstruction from a single image}

Reconstructing 3D object representations from a single image is an ill-posed problem, and it has clasically been tackled by baking in \emph{priors} about the shapes of objects being reconstructed. These shape priors are encoded using simple 3D primitives \cite{lawrence1963machine,nevatia1977description} or learned from a large set of repositories of 3D CAD models \cite{zia2013detailed,rock2015completing,bongsoo2015enriching}. There have also been approaches wherein large amount of training data is used to recover the shape of an object from a single image \cite{vicente2014reconstructing,kar2015category,krishna-icra2017,parkhiya2018constructing}.

Over the last $4-5$ years, research efforts have focused on reducing the amount of prior knowledge baked into the above approaches. For instance, Pixel2Mesh~\cite{wang2018pixel2mesh}, only assumes that the object being reconstructed has a spherical topology, and applies iterative deformations to an input spherical/ellipsoidal mesh to approximate a 3D object (mesh) in a supervised learning setting. While other approaches investigated pointcloud~\cite{fan2017point,groueix2018papier} and voxelgrid~\cite{choy20163d,rezende2016unsupervised,girdhar2016learning,wu2016learning,wu20153d} reconstruction,  the shapes produced by these approaches are often sparse, and miss out on high frequency detail. Recently, \emph{occupancy networks}~\cite{onet} were proposed, that treat reconstruction as a task of finding a decision boundary of an occupancy function, parameterized by a neural network.
We use 3D mesh representation as they provide closed surfaces that can be plugged in differential rasterization frameworks along with having less memory footprint. We use occupancy networks as they perform better and give high quality closed surfaces compared to other mesh reconstruction methodologies like Pixel2Mesh\cite{wang2018pixel2mesh} and AtlasNet\cite{groueix2018papier}.

\subsection{6-DoF pose recovery}

We also briefly review methods that \emph{only} aim to recover the 6-DoF pose of an object from an image, but not its shape.
Geometric approaches~\cite{RGB1, RGB2, RGB3} that rely on feature correspondences suffer drastic performance degradation with varation in viewpoints and occlusion. A few learning-based methods~\cite{pnp1, pnp2, pnp3} address this challenge by predicting 2D keypoints and then computing object poses using PnP techniques.

Another line of work uses RGB-D images~\cite{depth, posecnn, icp} with posthoc refinement (using Iterative Closest Point techniques). Recent methods~\cite{densefusion, P2Gnet} better exploit the complementary nature of color and depth information by locally fusing features. P2Gnet~\cite{P2Gnet} also uses object point clouds as priors and performs pose-guided point cloud generation in an end-to-end framework. To extend pose estimation to unseen object instances, \cite{nocs} introduces a Normalized Object Coordinate Space (NOCS) representing all possible object instances within a category. Alternatively, Wang \etal~\cite{wang20196} learn anchor keypoints representing the motion of an object instance. Our proposed framework instead leverages advances in differentiable rasterization~\cite{liusoft} to achieve accurate pose estimation and shape reconstruction.

\subsection{Object based SLAM}

Recent advances in SLAM methodologies and deep learning have enabled incorporating objects in SLAM frameworks thus enhancing their performance. SLAM++~\cite{salas2013slam++} uses a database of 3D scanned objects using depth cameras and performs an object-level slam by using an adapted KinectFusion~\cite{newcombe2011kinectfusion} method. A few other SLAM systems~\cite{dong2017visual,pillai2015monocular,dame2013dense} use point-based features along with objects detected in the scene to construct richer maps. 
In these approaches objects are represented as spheres\cite{frost2018recovering,sucar2018bayesian}, 3D ellipsoids\cite{rubino20173d} and quadrics\cite{nicholson2018quadricslam} which improved the scale drift in SLAM. 
Multiple efforts have explored other object representations~\cite{frost2018recovering,parkhiya2018constructing,sucar2018bayesian,rubino20173d,quadricslam,cubeslam} as well.
In \rrc{}, we estimate triangle meshes from monocular images, and leverage them for a subset of SLAM tasks, such as for object-centric egomotion estimation.

\subsection{Differentiable rendering}

In order to generate an image of the 3D mesh, the vertices are back projected onto the screen space and the image is generated through grid sampling, this process is called \textit{rendering}. The latter operation of grid sampling called \textit{rasterization} is not differentiable and hence it is difficult to incorporate in optimization tasks. 
Recently  Loper and Black \cite{opendr} introduce an approximate differentiable renderer which generates derivatives from projected pixels to the 3D parameters.
Kato \textit{et al}\cite{kato2018neural} also proposed a differentiable renderer which enables back-propagation by using an approximate gradient for rasterization. Liu \textit{et al}\cite{liusoft} views rendering as an aggregation function that fuses the probabilistic contributions of all mesh triangles with respect to the rendered pixels.
These differentiable rendering techniques have opened new frontiers in optimizing texture, lighting, object shapes and poses. We use differentiable renderer by \cite{kato2018neural} in our proposed framework (\rrc) to optimize for the pose of the object and show improvements in 6-DOF pose recovery compared to prior methods.

\section{Reconstruct, Render, and Backprop}
\label{sec:pipeline}
In this section we discuss the various building blocks of the framework and also expand on how they are incorporated into the pipeline. But before going further we would like to formulate the problem as:

\emph{Given an image $\mcal{I} \in \mathbb{R}^{H \times W \times C}$ ($H$ Height, $W$ Width and $C$ Channel) containing an object-of-interest, we aim to estimate}

\begin{compactitem}
\item \emph{The $6$-DoF pose $T \in SE(3)$ of the object-of-interest, relative to the camera coordinate frame.}
\item \emph{A mesh $\mcal{M}$ describing the 3D shape of the object of interest. In this work, we use triangle meshes to represent shape, i.e., $\mcal{M} = \left( \mcal{V}, \mcal{F} \right)$, where $\mcal{V}$ is a set of $N$ vertices of the mesh ($\mcal{V} \in \mathbb{R}^{N \times 3}$) and $\mcal{F}$ is a set of $F$ faces of the mesh ($\mcal{F} \in \mathbb{R}^{F \times 3}$)}
\end{compactitem}

\subsection{Occupancy Network}
Occupancy networks \cite{onet} view signed distance representations (SDF) proposed by \cite{curless1996volumetric} as a \emph{function space} and aim to learn a decision boundary that separates points that are on a surface from ones that aren't. Simply put, an \emph{occupancy function} $\mcal{O}(\mb{x}; \theta): \mathbb{R}^3 \mapsto \left[0, 1\right]$ takes a query location $\mb{x}$ in 3D, and produces a scalar $p_\theta(\mb{x})$ that can be loosely interpreted as the probability that $\mb{x}$ is occupied, i.e., that $x$ lies on the surface of the object-of-interest.\footnote{Note that the output $p_\theta(\mb{x})$ is usually a softmax activation, and has severe practical consequences when interpreted as a ``probability" \cite{gal2016dropout}.  For the scope of this work, however, we adopt this notion, as done in \cite{onet}.}
Occupancy networks are neural networks that are tasked with learning the occupancy function in a supervised setup. Often, occupancy networks use more context than just the 3D query point $\mb{x}$. Examples of such context would be convnet features $\phi(\mb{u})$ at a query location in an image $\mb{u}$.

\begin{figure}[t]
\begin{center}
   \includegraphics[width=0.99\linewidth]{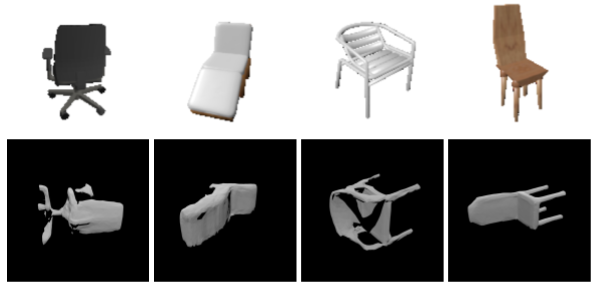}
\end{center}
   \caption{Qualitative results from the Occupancy Network. As can be seen the poses of output meshes are in a canonical frame which significantly differ from the ground truth pose of the object.}
\label{fig:onet_output}
\end{figure}

A discretized grid is taken on which the probability of a point lying inside or on the surface of the mesh is obtained using occupancy networks. This grid is then passed through a Multiresolution Isosurface Extraction algorithm \cite{onet} which extracts a mesh from this grid of occupancies. Figure \ref{fig:onet_output} illustrates an output of the occupancy network in the canonical frame.

\begin{figure*}[ht]
\begin{center}
\includegraphics[width=0.99\linewidth]{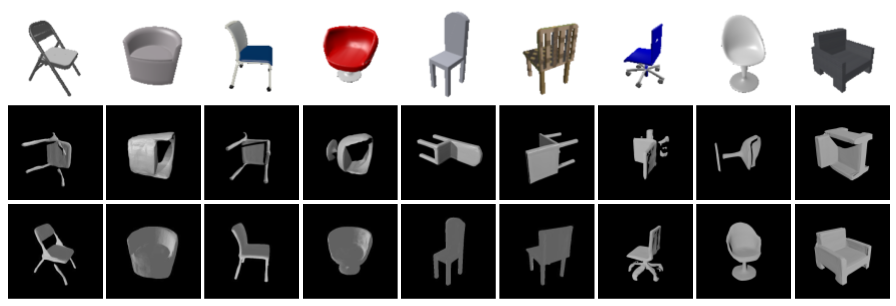}
\end{center}
   \caption{\textbf{First row}: Input images, \textbf{Second row}: Output of occupancy network in the canonical frame, \textbf{Third row}: Output of our proposed framework(\rrc) in the camera frame. \rrc{} takes the reference input image and converts it into a mesh in canonical frame. The output pose is then obtained from pose optimization using the differentiable renderer.}
\label{fig:qualitative}
\end{figure*}

\subsection{Viewpoint Network}
Occupancy networks output the meshes in a canonical frame, which is not useful to obtain the pose of the object. Hence we use RenderForCNN \cite{r4cnn} to generate a viewpoint estimate of the camera. This gives us azimuth, elevation and in-plane rotation angles of the camera with respect to the object, but for our purpose we use only the azimuth and elevation angles. This network is trained on a large set of synthetic dataset with objects from ShapeNet\cite{chang2015shapenet} rendered on random backgrounds of SUN397 dataset\cite{patterson2012sun}. 

\subsection{Differentiable Neural Renderer}

We use the Neural 3D mesh renderer by Kato et al\cite{kato2018neural} for the downstream task of recovering accurate pose of the mesh with respect to the camera. The renderer converts the rasterization operation of a mesh into 2D images differentiable, and hence enables us to solve for the pose of the mesh in the camera frame. We use the viewpoint estimate by the viewpoint network as the initialization and the known camera height and then iteratively optimize for the pose of  object with respect to  camera.

\textit{Inferring scale of the object}:
The 3D mesh obtained from the occupancy networks is in a different scale as the ground truth. Thus before optimizing for the pose we need to correct the scale of the mesh.
Consider the rotation matrix $R$ and the translation estimation $c=[x,y,z]$ which we obtain from the viewpoint estimator and the given camera height.
To get the correct scale, we make use of the bounding box of the object in the image. 
 Let \{$X_i^o$\} be the set of vertices of the mesh obtained from occupancy networks which will be in an arbitrary scale. The mesh in the camera frame will then be:
$$
X_i^c = R^T(X_i^o - c)
$$
where \{$X_i^c$\} is the set of mesh vertices in the camera frame. Let $K$ be the camera calibration matrix. We then back project the points on to the image using $K$ by using $x_i=KX_i^c$ where \{$x_i$\} is the set image pixels corresponding to $X_i^c$. We then get the bounding box of the reprojected object in the new image. This bounding box will be of a different size from the original image since the meshes are in different scales. We then recursively change the size of the mesh so that the reprojected object fits the bounding box, \(i.e\) we scale $X_i^c$. We can do this since we assume that the object will be vertically placed on the ground. This gives us the occupancy network mesh in the ground truth scale.

Let the modified rasterization operation of the differentiable renderer be \(g\{.\}\) which takes the 2D vertices \(\{x_i\}\) and transforms them into a differentiable space \(\{x_i^{'}\}\) \(\in\mathbb{R}^2\):
$$
g(x_i,f_k) = \{x_i^{'}, f_k\}
$$
where \(f^k \in \mcal{F}\).
We then minimize the Huber loss between the silhouette of the object in the rasterized image and the that in the reference image by simply subtracting the two images, thus iteratively solving for $R$ and $c$.

\section{Experiments and Results}
\label{sec:results}
In this section we present the experimental results of our proposed framework in terms of pose recovery. We use the dataset provided by Choy et al\cite{choy20163d} for the pose and localization errors. This dataset contains ShapeNet objects\cite{chang2015shapenet} rendered with varying viewpoints. Since we need to find the translation from a single image, we make use of the assumption that we know the height of the camera. We evaluate our approach on the following metrics: azimuth angle, elevation angle and 3D localization errors compared to the ground truth. We also include 3D bounding box error proposed by \cite{Wang_2019_CVPR} between the ground truth mesh and our mesh. In figure \ref{fig:qualitative}, we show the qualitative results which reveal dense mesh reconstruction and accurate pose recovery from single images.

We benchmark our method against Parv \textit{et al} \cite{parkhiya2018constructing} which is one such method that obtains 6 DoF pose of object from a single image, and show that our proposed method outperforms the former in nearly all the metrics. \cite{parkhiya2018constructing} uses keypoints detected from the hourglass model \cite{newell2016stacked} and constructs category specific wireframe models. Then they optimize over the pose and localization of the object whilst fitting the wireframe models to the keypoints detected in images. In this process they obtain the pose and localization of the objects in camera frame with which we compare our results. In order to calculate the 3D IOU, we find the 3D bounding box of the mesh using the vertices. We also compare the pose of \rrc{} with that of the mesh from occupancy network(OCC-NET). This error is expected to be high since the occupany network outputs mesh in the canonical frame. To give further insight into the efficacy of the differentiable renderer, we tabulate the pose error of the viewpoint estimation network(VIEW-NET)\cite{r4cnn}. We see that using the differentiable renderer improves the results from the viewpoint estimation network by a significant margin. All the results are tabulated in table \ref{tab:quantitative_result_table}.

\begin{figure}[ht!]
\begin{center}
   \includegraphics[width=0.95\linewidth]{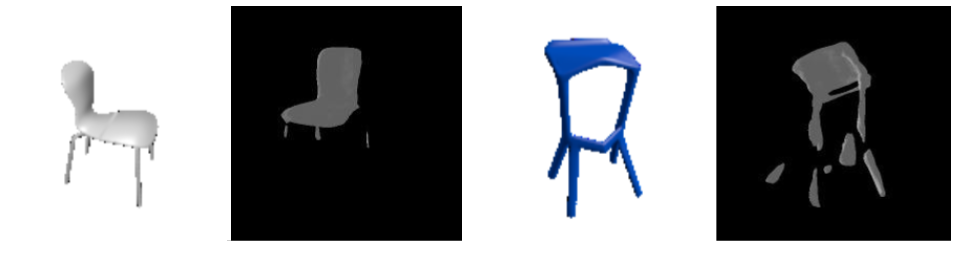}
\end{center}
  \caption{\textbf{Failure cases} The proposed method fails in cases where the OCC-NET mesh is partially or poorly constructed. As can been seen \rrc{} outputs a wrong pose. This usually happens when there are thin or disconnected structures in the 3D mesh.}
  \label{fig:failure}
\end{figure}

\begin{table}[h]
\begin{center}
\resizebox{1\linewidth}{!}
{
\begin{tabular}{|l|c|c|c|c|}
\hline
Method & 3D IOU & \makecell{Azimuth \\ (degrees)} & \makecell{Elevation \\ (degrees)} & \makecell{Translation \\ (meters)} \\
\hline\hline
OCC-NET & -- & 98.592 & 27.235 & --\\
VIEW-NET &  -- & 61.174 & 17.604 & --\\
Parv et.al \cite{parkhiya2018constructing} & 0.0847 & \textbf{8.165} & 20.282 & 1.820 \\
Ours & \textbf{0.7795} & 10.793 & \textbf{5.561} & \textbf{0.634} \\
\hline
\end{tabular}
}
\caption{Quantitative evaluation: This table shows that our method outperforms \cite{parkhiya2018constructing} and the baselines established by OCC-NET and VIEW-NET by huge margins.}
\label{tab:quantitative_result_table}

\end{center}
\end{table}
\vspace{-0.5cm}
In figure \ref{fig:failure} we show few failure cases where our framework fails due to erroneous occupancy network output. It might happen in cases with very thin structures in the mesh or with multiple disconnected parts.

\subsection{Application}
In this section we showcase an application of the proposed pipeline in which we perform object-centric egomotion estimation using only objects and compare it with the camera trajectory from monocular ORB SLAM \cite{orb}. We get the poses of the camera with respect to the object using our pipeline and use the inter camera poses to obtain a trajectory. Object-centric egomotion implies that all the camera poses are defined in the reference frame of the object. We use the initialization of the viewpoint network only for the first frame and then use the optimized viewpoint of the current frame as the initialization for the next frame.

\begin{figure}[htbp]
\begin{center}
\includegraphics[width=\linewidth]{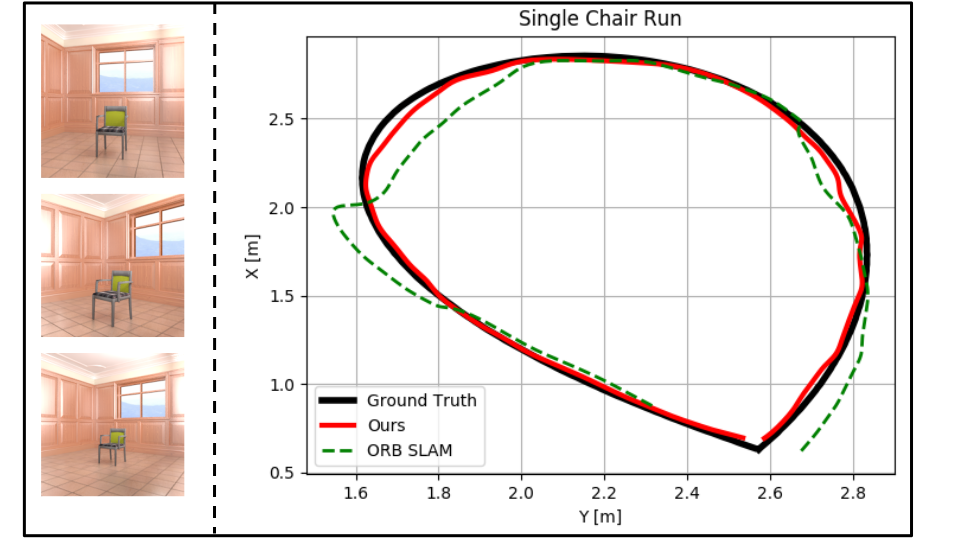}
\includegraphics[width=\linewidth]{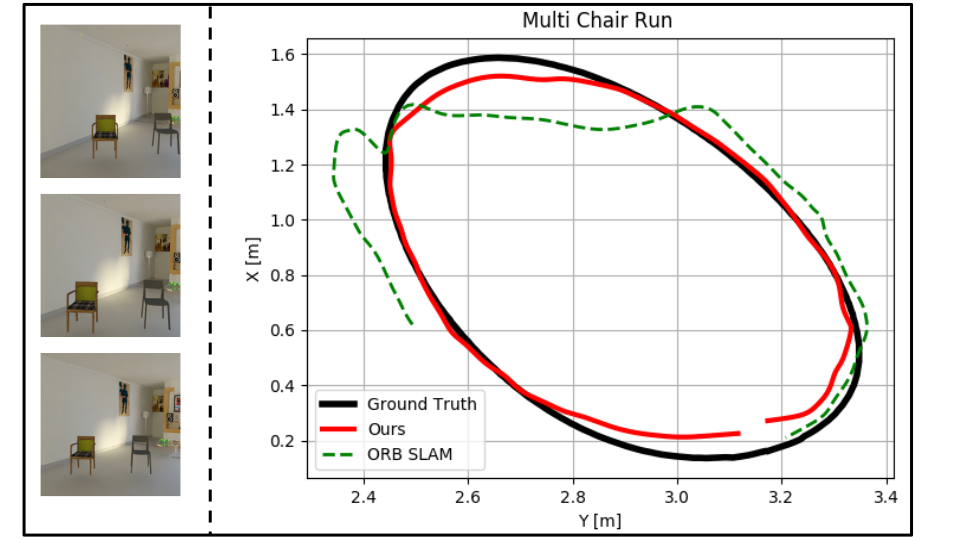}
  \caption{\textbf{Top}: Sample images and trajectory comparison of egomotion estimation using our proposed method with ORB Slam and the ground truth. This sequence uses a single chair for tracking. \textbf{Bottom}: A similar run with multiple chairs in the environment. As can be seen we show improved results compared to ORB SLAM.}
  \label{fig:slam_results}
  \end{center}
\end{figure}

For evaluating the results we use two synthetic runs (with chair object category) rendered in Blender. We assume that the ground-truth masks of the object (chairs in this case) are pre-computed. The first run contains the camera tracking a single chair. In the second run, we have 2 chairs and the camera moves in an elliptical path. We report the absolute trajectory error (ATE) and relative pose error (RPE) for both the runs. Refer to Table \ref{tab:slam_quantitative_results} and figure \ref{fig:slam_results} for quantitative and qualitative evaluations respectively.
It is notable that egomotion estimation using the proposed method is performed only using the objects and the known initial height of the camera. It does not require any other additional information or point features and performs comparable/better than the state of the art monocular ORB SLAM. This asserts the accuracy and robustness of our pipeline for the various downstream robotics task such as aiding the current state of the art SLAM algorithms in texture less environments, re-localization using camera in case of track loss, loop closing, etc. 

\begin{table}[ht!]
\begin{center}
\resizebox{0.9\linewidth}{!}
{
\begin{tabular}{|c|c|c|c|}\hline
Category & Comparison & ATE(m) & RPE(m) \\ \hline \hline
\multirow{2}{*}{\rule{0pt}{2ex}Single Chair Run} & ORB slam & 0.0955 & 1.0138 \\
& Ours & \textbf{0.0256} & \textbf{0.0505} \\ \hline
\multirow{2}{*}{\rule{0pt}{2ex}Multi Chair Run} &  ORB slam & 0.1122 & 0.8260 \\
& Ours & \textbf{0.0582} & \textbf{0.2116} \\ \hline
\end{tabular}
}
\caption{Quantitative evaluation: Absolute Trajectory and Relative pose error for single and multi chair run. Results indicate that our proposed method outperforms ORB slam in both the metrics.} 
\label{tab:slam_quantitative_results}
\end{center}
\end{table}

\section{Conclusions}
\label{sec:conclusion}

In this paper we have presented an approach for dense shape and pose estimation from a single image. It is built on the recently developed idea of differentiable rasterization which enables approximate gradient descent over the discrete operation of rasterization. Although the proposed framework does not perform real time, it is suitable for offline applications where accuracy is a bigger constraint. This is the first method of its kind which constructs a dense mesh reconstruction as well as estimates the pose of the objects from a single monocular image. We also demonstrate results on a monocular object-centric egomotion estimation setup using only 3D objects as features and showcase the application of this system for robotics related tasks. For our optimization task we use the ground truth masks of the objects which is one of the caveats of this work. Therefore future works could focus on improvements in obtaining accurate 2D object segmentation masks from real world images. This would be a stepping stone in this domain and aid the implementation in real world applications.

\section*{Acknowledgment}
The authors acknowledge the support and funding from Kohli Center for Intelligent Systems (KCIS), IIIT Hyderabad for this work. We also acknowledge the help of Junaid Ahmed Ansari, Gourav Kumar, Udit Singh Parihar, Aakash KT, Chanakya Vishal, Ashish Kubade and Kaustubh Mani for their timely assistance.

{\small
\bibliographystyle{ieee_fullname}
\bibliography{main}
}

\end{document}